\documentclass{article}

\usepackage{arxiv}

\usepackage[utf8]{inputenc} % allow utf-8 input
\usepackage[T1]{fontenc}    % use 8-bit T1 fonts
\usepackage{hyperref}       % hyperlinks
\usepackage{url}            % simple URL typesetting
\usepackage{booktabs}       % professional-quality tables
\usepackage{amsmath,amssymb}
\usepackage{amsfonts}       % blackboard math symbols
\usepackage{nicefrac}       % compact symbols for 1/2, etc.
\usepackage{microtype}      % microtypography
\usepackage{cleveref}       % smart cross-referencing
\usepackage{lipsum}         % Can be removed after putting your text content
\usepackage{graphicx}
\usepackage{natbib}
\usepackage{doi}

\usepackage{subcaption}
\usepackage{xurl}

\title{Benevolent Dictators? \\ On LLM Agent Behavior in Dictator Games}

\date{}

\newif\ifuniqueAffiliation
\ifuniqueAffiliation % Standard variant of author block
\author{\href{https://orcid.org/0009-0009-2271-2125}{\includegraphics[scale=0.06]{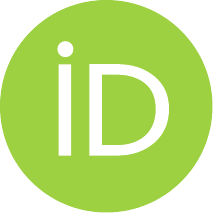}\hspace{1mm}Andreas~Einwiller}\\
	Professorship of Applied Machine Learning\\
	University of Passau\\
    Passau, Germany\\
	\texttt{andreas.einwiller@uni-passau.de}\\
	\And
	\href{https://orcid.org/0000-0003-4171-0597}{\includegraphics[scale=0.06]{orcid.pdf}\hspace{1mm}Kanishka~Ghosh~Dastidar}\\
	Chair of Distributed Information Systems\\
	University of Passau\\
    Passau, Germany\\
	\texttt{kanishka.ghoshdastidar@uni-passau.de}\\
    \And
	\href{https://orcid.org/0009-0004-4109-0468}{\includegraphics[scale=0.06]{orcid.pdf}\hspace{1mm}Artur~Romazanov}\\
	Computational Rhetoric and Natural Language Processing\\
	University of Passau\\
    Passau, Germany\\
	\texttt{artur.romazanov@uni-passau.de}\\
    \And
	\href{https://orcid.org/0000-0002-5901-9633}{\includegraphics[scale=0.06]{orcid.pdf}\hspace{1mm}Annette~Hautli-Janisz}\\
	Computational Rhetoric and Natural Language Processing\\
	University of Passau\\
    Passau, Germany\\
	\texttt{annette.hautli-janisz@uni-passau.de}\\
    \And
	\href{https://orcid.org/0000-0003-3566-5507}{\includegraphics[scale=0.06]{orcid.pdf}\hspace{1mm}Michael~Granitzer}\\
	Chair of Data Science\\
	University of Passau\\
    Passau, Germany\\
	\texttt{michael.granitzer@uni-passau.de}\\
    \And
	\href{https://orcid.org/0000-0001-7620-1376}{\includegraphics[scale=0.06]{orcid.pdf}\hspace{1mm}Florian~Lemmerich}\\
	Professorship of Applied Machine Learning\\
	University of Passau\\
    Passau, Germany\\
	\texttt{florian.lemmerich@uni-passau.de}\\
}
\else
\usepackage{authblk}

\newbox{\orcid}\sbox{\orcid}{\includegraphics[scale=0.06]{orcid.pdf}} 
\author[1]{%
    \\
	\href{https://orcid.org/0009-0009-2271-2125}{\usebox{\orcid}\hspace{1mm}Andreas~Einwiller\thanks{\texttt{andreas.einwiller@uni-passau.de}}}%
}
\author[1]{%
	\href{https://orcid.org/0000-0003-4171-0597}{\usebox{\orcid}\hspace{1mm}Kanishka~Ghosh~Dastidar\thanks{\texttt{kanishka.ghoshdastidar@uni-passau.de}}}%
}
\author[1]{%
	\href{https://orcid.org/0009-0004-4109-0468}{\usebox{\orcid}\hspace{1mm}Artur~Romazanov\thanks{\texttt{artur.romazanov@uni-passau.de}}}%
}
\author[1]{%
    \\
	\href{https://orcid.org/0000-0002-5901-9633}{\usebox{\orcid}\hspace{1mm}Annette~Hautli-Janisz\thanks{\texttt{annette.hautli-janisz@uni-passau.de}}}%
}
\author[1]{%
	\href{https://orcid.org/0000-0003-3566-5507}{\usebox{\orcid}\hspace{1mm}Michael~Granitzer\thanks{\texttt{michael.granitzer@uni-passau.de}}}%
}
\author[1]{%
	\href{https://orcid.org/0000-0001-7620-1376}{\usebox{\orcid}\hspace{1mm}Florian~Lemmerich\thanks{\texttt{florian.lemmerich@uni-passau.de}}}%
}
\affil[1]{University of Passau, Passau, Germany}
\fi

\renewcommand{\shorttitle}{Benevolent Dictators? On LLM Agent Behavior in Dictator Games}

\hypersetup{
pdftitle={Benevolent Dictators? On LLM Agent Behavior in Dictator Games},
pdfsubject={In behavioral sciences, experiments such as the ultimatum game are conducted to assess preferences for fairness or self-interest of study participants. In the dictator game, a simplified version of the ultimatum game where only one of two players makes a single decision, the dictator unilaterally decides how to split a fixed sum of money between themselves and the other player. Although recent studies have explored behavioral patterns of AI agents based on Large Language Models (LLMs) instructed to adopt different personas, we question the robustness of these results. In particular, many of these studies overlook the role of the system prompt—the underlying instructions that shape the model's behavior—and do not account for how sensitive results can be to slight changes in prompts. However, a robust baseline is essential when studying highly complex behavioral aspects of LLMs. To overcome previous limitations, we propose the LLM agent behavior study (LLM-ABS) framework to (i) explore how different system prompts influence model behavior, (ii) get more reliable insights into agent preferences by using neutral prompt variations, and (iii) analyze linguistic features in responses to open-ended instructions by LLM agents to better understand the reasoning behind their behavior. We found that agents often exhibit a strong preference for fairness, as well as a significant impact of the system prompt on their behavior. From a linguistic perspective, we identify that models express their responses differently.
Although prompt sensitivity remains a persistent challenge, our proposed framework demonstrates a robust foundation for LLM agent behavior studies.},
pdfauthor={Andreas~Einwiller, Kanishka~Ghosh~Dastidar, Artur~Romazanov, Annette~Hautli-Janisz, Michael~Granitzer, Florian~Lemmerich},
pdfkeywords={LLMs, Behavior assessment, Dictator game, Linguistic analysis},
}

\begin{document}
\maketitle

\begin{abstract}
In behavioral sciences, experiments such as the ultimatum game are conducted to assess preferences for fairness or self-interest of study participants.
In the dictator game, a simplified version of the ultimatum game where only one of two players makes a single decision, the dictator unilaterally decides how to split a fixed sum of money between themselves and the other player.
Although recent studies have explored behavioral patterns of AI agents based on Large Language Models (LLMs) instructed to adopt different personas, we question the robustness of these results.
In particular, many of these studies overlook the role of the system prompt—the underlying instructions that shape the model's behavior—and do not account for how sensitive results can be to slight changes in prompts.
However, a robust baseline is essential when studying highly complex behavioral aspects of LLMs.
To overcome previous limitations, we propose the LLM agent behavior study (LLM-ABS) framework to (i) explore how different system prompts influence model behavior, (ii) get more reliable insights into agent preferences by using neutral prompt variations, and (iii) analyze linguistic features in responses to open-ended instructions by LLM agents to better understand the reasoning behind their behavior.
We found that agents often exhibit a strong preference for fairness, as well as a significant impact of the system prompt on their behavior. From a linguistic perspective, we identify that models express their responses differently.
Although prompt sensitivity remains a persistent challenge, our proposed framework demonstrates a robust foundation for LLM agent behavior studies.
Our code artifacts are available at \url{https://github.com/andreaseinwiller/LLM-ABS}.
\end{abstract}

% keywords can be removed

\section{Introduction}
Given the rapid adoption of generative models in many areas of human interaction, we predict that LLMs will eventually evolve from passive text generators on demand to autonomous LLM-based agents involved in complex decision-making processes.
It is therefore crucial to study how these agents behave in different scenarios. Specifically, we need to understand whether they act rationally, strategically, or in human-like ways—and in the latter case, whether their behavior tends to be altruistic or selfish.
This behavioral characterization will be essential for predicting and governing their interactions in real-world scenarios.
An established method to study these social preferences in human subjects is behavioral games from game theory such as the \emph{dictator game}.
In the dictator game, a participant (the ``dictator'') receives a certain amount of money and unilaterally decides how much they keep for themselves and how much is transferred to another (powerless) recipient.
Since the recipient has no say in the outcome, any amount the dictator chooses to give reflects purely non-strategic behavior—such as generosity, fairness, or concern for the other person.

First works have transferred the dictator game to LLM agents by prompting the models to adopt different personas and split the money in different scenarios~\citep{ma2024can,brookins2023playing, filippas2024large}.
However, these scenarios and personas do not necessarily resemble everyday human-AI interactions and exclude potential systematic variations such as system prompts.
Similarly, models tend to be sensitive to slight user prompt variations, which has a significant effect on decision-making.
This is particularly relevant, as agentic decision-making can have (known) biases, but must be consistent.

To establish a strong baseline and study systematic variations as well as human-like behavior and consistency,  we propose the LLM agent behavior study (LLM-ABS) framework to (i) explore how different system prompts influence model behavior, (ii) get more reliable insights into agent preferences in terms of strategic behavior, altruism, and consistency by using neutral prompt variations, and (iii) analyze linguistic features in responses to open-ended instructions by LLM agents to better understand the reasoning behind their behavior.
We prompt several diverse, up-to-date LLM models with task descriptions from experiments on human behavior in dictator games. To obtain robust estimates of LLM agent behavior, we evaluate various system prompts and integrate multiple neutral variations for each user prompt.
Moreover, we assess the effect of changing the amount or unit and compare the results of our experiments with known findings on human behavior.

\section{Related Work}

\paragraph{Dictator Games in LLMs.}
Using games to study the behavior of humans has a long history in the social sciences.
The dictator game~\citep{forsythe1994fairness} in particular has been employed to study the role of fairness, altruism, and self-interest in human behavior.
According to a meta-study by~\citet{engel2011dictator}, humans on average give away about 28\% of the amount to be distributed with significant influence of the participant's demographic, anonymity, and other factors.

More recently, dictator games have also been employed to investigate the behavior of AI agents.
In this direction, early experiments have observed human-like behavior with tendencies towards fairness for ChatGPT 3.5~\citep{brookins2023playing}.
Similarly,~\citet{filippas2024large} confirmed that experiments with LLMs, particularly GTP 3.5 and GPT-4-preview, as subjects in such games can replicate results from studies with human participants~\citep{filippas2024large}.
\citet{mozikov2024eai} demonstrated that in such situations, LLMs also are impacted by emotion cues in games such as dictator and ultimatum games.
The SUVA framework (``State-Understanding-Value-Action") was proposed to analyze LLM-behavior in social contexts including the dictator game~\citep{leng2023llm}.
This approach focuses on understanding the reasoning process, e.g., including the chain-of-thought.
By contrast, the LLM-ABS framework introduced in this paper focuses on the robustness of results over a wide variety of recent models leveraging manually created neutral prompt variations. 

The work that we consider as closest to ours is by~\citet{ma2024can}.
It studies the behavior of LLMs prompted to mimick different \emph{personas} by setting demographic variables,
LLM temperature values, and personality traits.
However, in contrast to our work, they do not investigate the impact of different system prompts in a robustness-oriented setting. 

\paragraph{Prompt sensitivity.}
LLMs are highly sensitive to input prompts, expressed by the fact that minor variations in phrasing or structure can lead to significant changes in the generated output~\citep{razavi2025benchmarking}.
This highlights the problem: how can we mitigate the extent to which LLM responses are overly dependent on specific wordings or even punctuation?
\citet{zhu2024prompt, razavi2025benchmarking} have proposed the generation of variations to benchmark robustness.
However, we are rather critical of transferring this approach to applications other than benchmarking, as generating variations with LLMs requires setting the temperature hyperparameter, which is associated with substantial uncertainty regarding neutrality.

\section{Experiments}

To study the behavior of an LLM agent, the response $y$ is quantitatively evaluated on the input $x$.
In particular, a distinction is made between system prompts $x_{\mathrm{sys}}$ and user prompts $x_{\mathrm{usr}}$, whereby user prompts include template variables.
Template variables such as \texttt{<A>} and \texttt{<U>} in \texttt{"Assume, you have <A><U>. Decide how to distribute them between you and a participant."} enable substitutions with various amounts and units.
For example, $x_{\mathrm{usr}}[A=\text{"10"},\,U=\text{"\$"}]$ transforms \texttt{<A><U>} into \texttt{10\$}.

This paper focuses on exploring the effect of system prompts, amounts, and units based on $8$ variations for each variable. For the underlying user prompt, $10$ neutral instructions, including the template variable amount and unit, were created manually to mitigate prompt sensitivity.
To evaluate the effect of one variable based on these $8$ variations, the other variables remain fixed.
Thus, when evaluating system prompts, the amount and unit variables in $x_{\mathrm{usr}}$ are fixed.
We have chosen the following default values: $x_{\mathrm{sys}}[\texttt{"You are a helpful assistant."}]$, $x_{\mathrm{usr}}[A=\text{"10"}]$, and $x_{\mathrm{usr}}[U=\text{"\$"}]$.

System prompt variations include an empty and three very short default system prompts, as well as four more realistic system prompts of DeepSeek\footnote{\tiny\url{https://github.com/LouisShark/chatgpt_system_prompt/blob/main/prompts/official-product/deepseek/R1.md}}, Claude\footnote{\tiny\url{https://github.com/LouisShark/chatgpt_system_prompt/blob/main/prompts/official-product/claude/Claude3.md}}, Grok\footnote{\tiny\url{https://github.com/LouisShark/chatgpt_system_prompt/blob/main/prompts/official-product/xai/Grok2.md}}, and Gemini\footnote{\tiny\url{https://github.com/LouisShark/chatgpt_system_prompt/blob/main/prompts/official-product/google/gemini-2-5pro-20250421.md}}.
Variables inside these system prompts, such as the current date, used to specify the knowledge cutoff, were removed.
The $8$ amount variations cover low and high discrete, special, and continuous numbers with $10$ and $20$ decimal places.
Unit variations include different currencies and representations of such, as well as \texttt{"liters of water"} and \texttt{"kilograms of rice"}.

\begin{figure}
    \centering
    \includegraphics{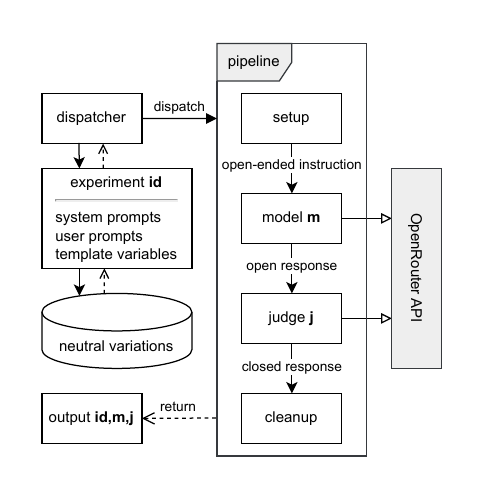}
    \caption{Component overview of the LLM agent behavior study (LLM-ABS) framework. Metadata related to each computation is collected throughout every stage of the pipeline.}
    \label{fig:framework}
\end{figure}

Experiments are repeated $10$ times for each of the $10$ manually created neutral user prompt instructions, resulting in up to $100$ observations.
We used the default temperature of $1.0$, as defined by OpenRouter\footnote{\tiny\url{https://openrouter.ai/docs/api-reference/parameters\#temperature}}, to simulate a realistic user setting and enable output variability ($\text{temperature} > 0$).
Figure~\ref{fig:framework} illustrates precisely how each input is processed within the pipeline of our proposed LLM agent behavior study (LLM-ABS) framework.
Table~\ref{tab:model-selection} shows the diverse model selection, which allows for a comprehensive evaluation across dimensions such as architecture and scale.
Each model generates an open(-ended) response based on the experiment input.
Generating an open response enables us to not only capture the decision made by an LLM, but also allows us to capture the reasoning which can be analyzed using linguistic methods.
This open response is reduced to a closed(-form) response using an LLM-as-a-Judge approach.
The closed response is instructed as valid JSON and encapsulates how much the agent kept, how much was distributed to the other participant, or whether the open response was classified as a refusal.
In order to simplify the pipeline for this initial paper, implementing an ensemble of judges was discarded.
Thus, each model is its own judge in the consecutive reduction step.
LLM queries were computed using the OpenRouter API\footnote{\tiny\url{https://openrouter.ai/}}.
Given each model is its own judge, up to $100$ observations were computed for each experiment and model combination.
This is an upper limit, as open responses can contain refusals or the closed response may contain an invalid format or values, which results in them being filtered out for the evaluation.

\begin{table}
    \centering
    \caption{Model selection and alias mapping.}
    \begin{tabular}{@{}llll@{}} % @{} removes space form vertical edges
        \toprule
        \textbf{Model} & \textbf{Alias} & \textbf{Type} & \textbf{Company} \\
        \midrule
        deepseek-chat-v3-0324 & DeepSeek & Text-only & DeepSeek \\
        claude-3-haiku & Claude & Multimodal & Anthropic \\
        grok-3-mini-beta & Grok & Text-only & xAI \\
        gemini-2.0-flash-001 & Gemini & Multimodal & Google \\
        gpt-4.1-mini & Gpt & Multimodal & OpenAI \\
        llama-3-70b-instruct & Llama & Text-only & Meta LLaMA \\
        qwen-2.5-7b-instruct & Qwen & Text-only & Qwen \\
        mixtral-8x7b-instruct & Mixtral & Text-only & Mistral AI \\
        \bottomrule
    \end{tabular}
    \label{tab:model-selection}
\end{table}

\section{Results}

\begin{figure*}
    \centering
    
    \begin{subfigure}{\textwidth}
        \begin{minipage}[c]{0.04\textwidth}
            \rotatebox{90}{\parbox{4cm}{\centering\small \textbf{(a) System prompt}}}
        \end{minipage}
        \begin{minipage}[c]{0.95\textwidth}
            \includegraphics[width=\textwidth]{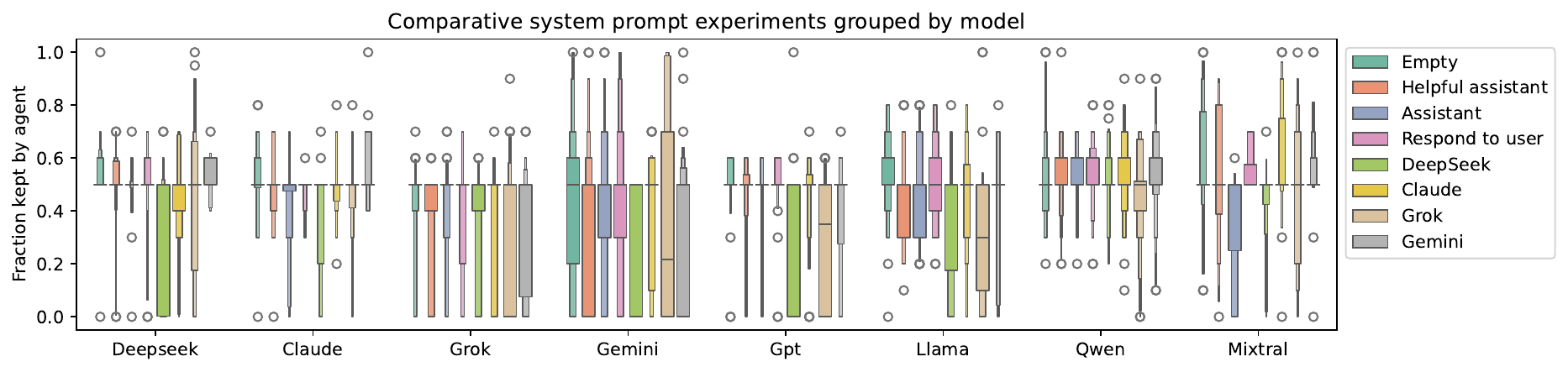}
        \end{minipage}
    \end{subfigure}

    \begin{subfigure}{\textwidth}
        \begin{minipage}[c]{0.04\textwidth}
            \rotatebox{90}{\parbox{4cm}{\centering\small \textbf{(b) Amount}}}
        \end{minipage}
        \begin{minipage}[c]{0.95\textwidth}
            \includegraphics[width=\textwidth]{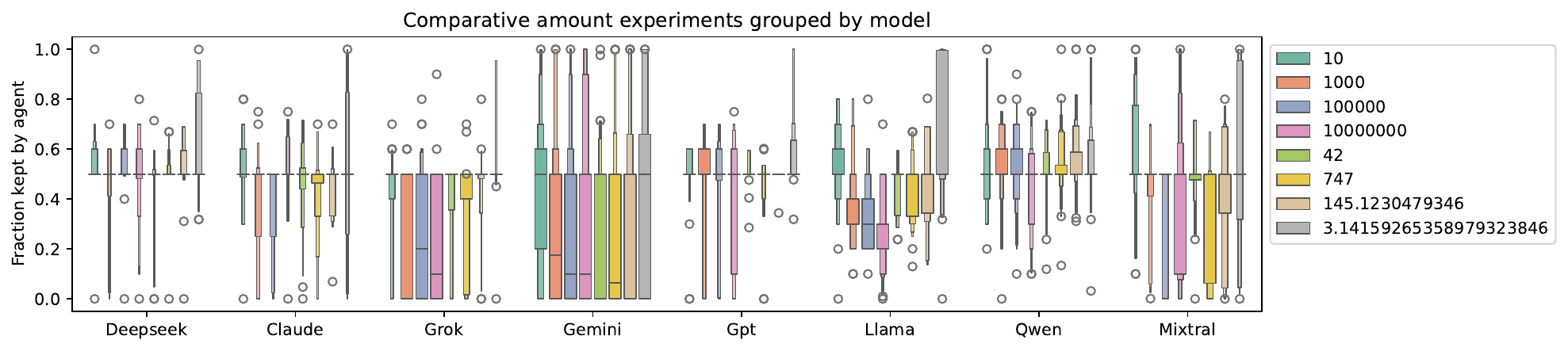}
        \end{minipage}
    \end{subfigure}

    \begin{subfigure}{\textwidth}
        \begin{minipage}[c]{0.04\textwidth}
            \rotatebox{90}{\parbox{4cm}{\centering\small \textbf{(c) Unit}}}
        \end{minipage}
        \begin{minipage}[c]{0.95\textwidth}
            \includegraphics[width=\textwidth]{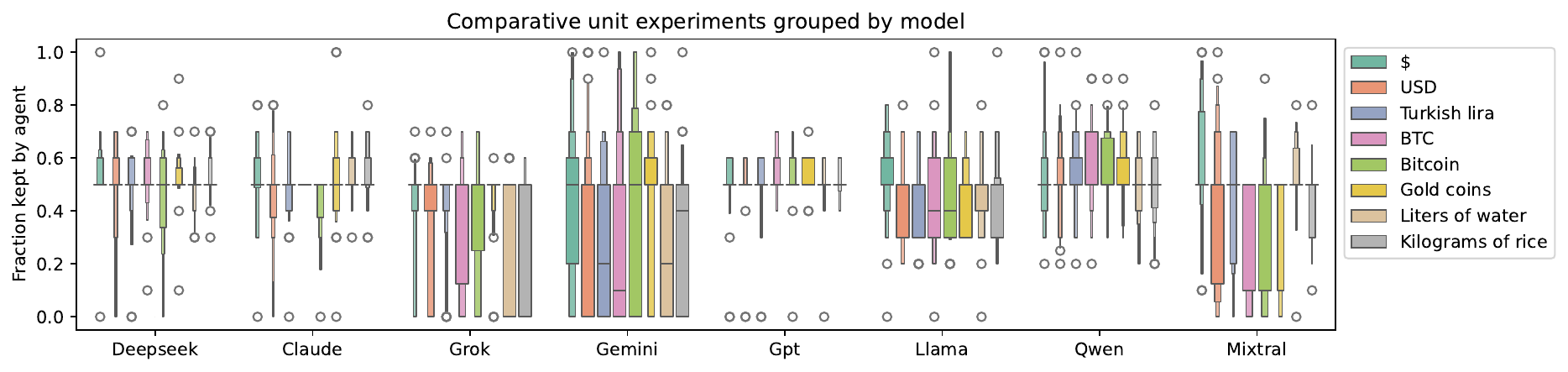}
        \end{minipage}
    \end{subfigure}
    
    \caption{Distribution of endowment split choices across models in response to variations in (a) system prompt, (b) amount, and (c) unit. Each letter-value plot shows up to $100$ observations, based on $10$ neutral user prompt variants combined with $10$ repetitions. Outliers, marked as circles, represent the most extreme $10\%$ of observations.}
    \label{fig:joint-comparative-results}
\end{figure*}

We present the results of our experiments using letter-value plots in Figure~\ref{fig:joint-comparative-results}.
This visualization offers a more detailed view of the data distribution compared to classic boxplots, as it can represent more than just the $25\%$, $50\%$, and $75\%$ quartiles ($Q_1$, $Q_2$, and $Q_3$).
For instance, if all data points within the interquartile range are identical, i.e., $\forall{x, y} \in [Q_1, Q_3], x = y$, then $Q_1 = Q_3$, and the interquartile range ($IQR$) becomes zero.
In such cases, $Q_1$, $Q_2$, and $Q_3$ coincide with the lower and upper whiskers, which may cause up to $50\%$ of the data to be classified as outliers.
To overcome this limitation, as noted by~\citet{letter-value-plot}, we used the seaborn.boxenplot\footnote{\tiny\url{https://seaborn.pydata.org/generated/seaborn.boxenplot.html}} implementation, setting the outlier proportion to $10\%$. 

\subsection{Behavioral Tendencies of LLM Agents}
The three graphs in Figure~\ref{fig:joint-comparative-results} show the retained proportion of the agent under varying configurations of system prompts, endowment amounts, and unit types.
The resulting insights into their behavioral tendencies can be analyzed along different dimensions:

\paragraph{Impact of System Prompt.}
Understanding how system-level instructions shape the values and decision-making of an agent is crucial.
Overall, shorter and seemingly neutral system prompts—such as `You are a helpful assistant.' or `You are an assistant.'—tend to produce only minor changes in the distribution of the fraction of money kept by the agent, with few exceptions (Mixtral).
In contrast, more elaborate system prompts, such as the one used by DeepSeek, appear to induce a marked shift toward generosity in several agents, including DeepSeek, Gemini, Gpt, and Llama.
Similarly, the system prompt associated with Grok also leads to substantial increases in generosity for Gemini, Gpt, and Llama.
By contrast, the system prompt used by Claude does not lead to significant behavioral changes across the agents, suggesting a comparatively weaker framing effect.
These results suggest that the content and framing of system-level instructions can significantly influence agent behavior.
However, this effect differs across agents, with some agents such as Claude or Llama not significantly altering their splits across different system prompts.

\paragraph{Altruism vs Self-interest.} 
The plots show that the central tendency of most agents is a $50$/$50$ split of the endowment amount, regardless of the given system prompt, amount or unit.
This is a far more altruistic division than the  economically rational (self-interested) choice, which would be to keep everything.
While studies of human behavior in the dictator game also do not show complete self-interest, humans retain on average more than a two-third share for themselves~\citep{engel2011dictator}.
In our experiments, we observe that for LLM agents, when deviations from the fair split do occur, they tend to lean toward even greater altruism, with some, such as Grok, Gemini, and Llama, frequently choosing to give away the majority of the endowment.
An exception to this pattern is Qwen, which displays a more frequent inclination toward self-interest compared to the other agents.
Possible explanations for these tendencies in agent behaviour are numerous and difficult to isolate.
These include biases in training and alignment, absence of self-interest or goals or the over-reliance on fairness as a safe or normative answer.

\paragraph{Consistency.}
We derive insights about the consistency of each agent's behavior when exposed to controlled perturbations in input that vary the task context without altering the agent's underlying identity or persona.
These include neutral variations of the user prompt and changing the amount or unit.
On one hand, such robustness is desirable in systems that must behave consistently across noisy or ambiguous user input.
However, we should also note that changing the amounts and units results in a task that is not exactly equivalent.
This is reflected in human responses where generosity tends to decrease with increasing amounts~\citep{engel2011dictator}. Notably, we observe that DeepSeek and Claude show very little deviations from a 50/50 split across all our settings.
On the other hand, Gemini, Grok, and to an extent, Mixtral, display much more erratic behavior. For Gemini, the proportion of the amount it chooses to retain varies starkly based on subtle semantic changes in the user prompt as well as to changes in the amount or unit.
In fact, in contrast to humans, the central tendency shifts strongly towards increased generosity for larger amounts of money. An interesting finding is that Gemini is far more altruistic when the unit is presented as `BTC' compared to `Bitcoin'.
This shows that seemingly small wording changes can produce significantly different behavior—a key concern for deploying LLMs in socially sensitive or decision-making roles.

\subsection{Linguistic Properties of the Responses}
In order to shed more light on how the linguistic structures of the responses differ, we analyze the responses in terms of their frequency of epistemic markers and the occurrence of discourse markers.
The former signals the commitment uttered in the responses, for instance through expressions such as `potentially', `definitely', `I think' and `I believe', a statement conveys different confidence of the agent.
To this end we adopt an empirically-driven approach and derive a set of epistemic markers from a corpus of argumentative discourse~\citep{QT30}, which we extract from the open responses. 
The use of discourse markers approximates the amount of justifications offered for a decision, measured by the frequency of discourse markers (e.g., `therefore', `because') and if-clauses for weighing and judging the options.
Here we again base ourselves on dictionaries, this time extracted from the Penn Discourse Treebank~\citep{pdtb}, where those markers and their function are annotated. 

We see that DeepSeek behaves slightly different compared to all other models. In particular the responses where the system suggests a fair split exhibit a higher degree and variation of discourse markers, showing that in these cases the model explicitly justifies why this decision was taken more than in the cases of favoring one of the sides.
This holds independent of the system prompt that was used. Gpt, in line with the other models, does not offer a large degree of justifications. 
Regarding epistemic markers, the fair distribution of money again yields a slightly different pattern in terms of the commitment structure.
DeepSeek stands out in comparison to the other models in that it generates a larger distribution of epistemic structures, in contrast to Gpt and Mixtral, which generate a smaller share of epistemic markers.

\section{Conclusion}
This paper presents LLM-ABS, a robust experimental framework designed to minimize prompt sensitivity when conducting LLM agent behavior studies.
Based on a series of dictator game experiments, we examine how agent behavior varies with changing system prompts, amounts, and units.
Our experiments reveal that shorter default system prompts exert only a small effect on LLM agents.
More elaborate system prompts, such as those from DeepSeek and Grok, induce a significant shift towards generous behavior across multiple agents.
We observe a central tendency towards an equal $50$/$50$ split, demonstrating greater altruism compared to human behavior.
Agents such as Grok, Gemini, and Llama frequently choose to give away the majority of the endowment.
Qwen, on the other hand, displays more self-interest, which could provide information related to bias in training and alignment.
While DeepSeek and Claude remain mostly stable across experiments, others respond with much more erratic behavior.
Our linguistic analysis reveals that DeepSeek exhibits a more varied discourse and a broader range of epistemic structures when justifying a fair split, indicating deeper reasoning.

\bibliographystyle{unsrtnat}
\bibliography{references}

\end{document}